# Brain-like combination of feedforward and recurrent network components achieves prototype extraction and robust pattern recognition

Naresh Balaji Ravichandran[1][0000-0001-7944-4226], Anders Lansner[1,2][0000-0002-2358-7815], Pawel Herman[1][0000-0001-6553-823X]

[1] KTH Royal Institute of Technology, Stockholm 100 44
[2] Stockholm University, Stockholm 106 91
{nbrav, ala, paherman}@kth.se

**Abstract.** Associative memory has been a prominent candidate for the computation performed by the massively recurrent neocortical networks. Attractor networks implementing associative memory have offered mechanistic explanation for many cognitive phenomena. However, attractor memory models are typically trained using orthogonal or random patterns to avoid interference between memories, which makes them unfeasible for naturally occurring complex correlated stimuli like images. We approach this problem by combining a recurrent attractor network with a feedforward network that learns distributed representations using an unsupervised Hebbian-Bayesian learning rule. The resulting network model incorporates many known biological properties: unsupervised learning, Hebbian plasticity, sparse distributed activations, sparse connectivity, columnar and laminar cortical architecture, etc. We evaluate the synergistic effects of the feedforward and recurrent network components in complex pattern recognition tasks on the MNIST handwritten digits dataset. We demonstrate that the recurrent attractor component implements associative memory when trained on the feedforward-driven internal (hidden) representations. The associative memory is also shown to perform prototype extraction from the training data and make the representations robust to severely distorted input. We argue that several aspects of the proposed integration of feedforward and recurrent computations are particularly attractive from a machine learning perspective.

**Keywords:** attractor; associative memory; unsupervised learning; Hebbian learning; recurrent networks; feedforward networks; brain-like computing.

## 1    Introduction

Recurrent networks are ubiquitous in the brain and constitute a particularly prominent feature of the neocortex [1–3]. Yet, it is not clear what function such recurrence lends to cortical information processing. One popular hypothesis is that recurrent cortical networks perform associative memory, where assemblies of coactive neurons (cell assembly) act as internal mental representation of memory objects [4–6]. Several theoretical and computational studies have shown that recurrently connected neuron-like binary units with symmetric connectivity can implement associative attractor memory: the network is guaranteed to converge to attractor states corresponding to local energy minima [4]. Learning memories in such networks typically follows Hebbian synaptic plasticity, i.e., the synaptic connections between units are strengthened when they are coactive (and weakened otherwise). Attractor networks provide a rich source of network dynamics and have produced several important network models of the neocortex that explain complex cognitive functions [7–9].

Typically, attractor networks are trained with artificially generated orthogonal or random patterns, since overlapping (non-orthogonal) inputs cause interference between memories, so-called crosstalk, and lead to the emergence of spurious memories [10]. Consequently, attractor networks have not really been combined with high-dimensional correlated input and the problem of extracting suitable representations from real-world data has not received much attention in the context of associative memory. It is hypothesized that desirable neural representations in the brain are extracted from sensory input by feedforward cortical pathways. This biological inspiration has been loosely adopted in deep neural networks (DNN) developed for pattern recognition on complex datasets, e.g., natural images, videos, audio, natural languages. The main focus in the DNN community is on learning mechanisms to obtain high-dimensional distributed representations by extracting the underlying factors from the training data [11]. The success of DNNs has been predominantly attributed to the efficient use of the backprop algorithm that adjusts weights in the network to minimize a global error in supervised learning scenarios. Since it is hard to justify the biological plausibility of backprop, there has been growing interest in recent years on more brain plausible local plasticity rules for learning representations [12–17].

The work we present here demonstrates a step towards a brain-like integration of recurrent attractors networks with feedforward networks implementing representation learning. In our model, the recurrent attractors are trained on feedforward-driven representations and the network exhibits multiple associative memory capabilities such as prototype extraction, pattern completion and robustness to pattern distortions. The model incorporates several core biological details that are critical for computational functionality [18]: unsupervised learning, Hebbian plasticity, sparse connectivity, sparse distributed neural activations, columnar and laminar cortical organization. We show results from evaluating our new model on the MNIST handwritten digit dataset, a popular machine learning benchmark.

## 2 Model description

### 2.1 Model overview

Our network consists of an input layer (representing the data) and a hidden layer (Fig. 1A). The feedforward network component connects the input layer to the hidden layer and the recurrent component connects the hidden layer to itself. The organization of within the hidden layer can be described in terms of vertical and horizontal organization as discussed below.

The vertical organization of the hidden layer comprises minicolumns and hypercolumns, which derives from the discrete columnar organization of the neocortex of large mammalian brains [19, 20]. The cortical minicolumn (shown as cylinders in Fig. 1A) comprises around 80-100 tightly interconnected neurons having functionally similar response properties [19–21]. The minicolumns are arranged in larger hypercolumn modules (shown as squares enclosing the cylinders in Fig. 1B), defined by local competition between minicolumns through shared lateral inhibition [20, 22].

The hidden layer is also organized horizontally in term of laminae as derived from the laminar structures found through the depths of the cortical sheet. The basis for such laminar organization in the cortex is mainly neuroanatomical, differing in their cell types, cell densities, afferent, and efferent connectivity, etc. In our network, we model units in layer 4 (L4; granular layer) and layers 2/3 (L2/3; supragranular) (we do not model L1, L5 and L6). The L4 units receive the feedforward connections from the input layer (sparse connectivity). The L2/3 units receive recurrent connections from other L2/3 units in the hidden layer (full connectivity) and the L4 units within the same minicolumn. The extensive recurrent connectivity within the L2/3 units in the hidden layer implements associative memory function [5, 23].

The computations performed by the units are based on the Bayesian Confidence Propagation Neural Network model (BCPNN) [24–27]. BCPNN converts probabilistic inference (naïve Bayes) into neural and synaptic computation [24, 25, 27] and the BCPNN computations are applicable to both the recurrent and feedforward components.

## 2.2   Activation rule

When source units (indexed by $i$) send connections to a target unit (index by $j$), the activity propagation rule is:

$$s_j = b_j + \sum_i \pi_i w_{ij}, \tag{1}$$

$$\pi_j = \frac{e^{s_j}}{\sum_k e^{s_k}}, \tag{2}$$

where $s_j$ is the total input received by the $j$-th target unit, $b_j$ and $w_{ij}$, are the bias and weight parameters, respectively (Fig. 1B). The activation $\pi_j$ is calculated by a softmax non-linear activation function that implements a soft-winner-takes-all competition between the units within each hypercolumn from the same lamina [22].

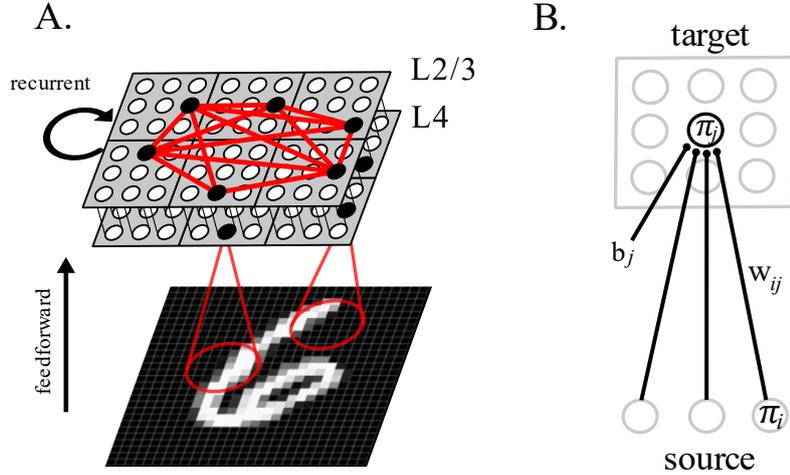

**Fig. 1. Network schematic. A.** The feedforward component connects the input layer to the hidden layer and the recurrent component connects the hidden layer to itself. The hidden layer is organized vertically in terms of minicolumns and hypercolumns, shown as cylinders and gray squares enclosing the cylinders respectively, and horizontally in terms of laminar layers L4 and L2/3, shown as gray sheet at the bottom and on top respectively. L4 units receive feedforward connections from the input layer and the L2/3 units receive recurrent connections from other L2/3 units (and from the L4 unit within the same minicolumn). In the schematic, we show $H_h = 6$ hypercolumns, each with $M_h = 9$ minicolumns. The activities of the units are indicated by black (1) and white (0) circles, and the red connection link indicate the cell assembly dynamically formed within the recurrent network component. **B.** In the BCPNN model, a target unit's activation is driven by source activations and determined by local competition within the target hypercolumn (gray square box). The BCPNN computations are universal and apply to both the feedforward and recurrent components.

### 2.3 Learning rule

The synaptic plasticity in our model is based on the incremental form of Hebbian-Bayesian learning [25, 27], where local traces of co-activation of pre- and post- synaptic units are used to compute Bayesian weights. The learning step involves incrementally updating three $p$-traces: probability of pre-synaptic activity, $p_i$, probability of post-synaptic activity, $p_j$, and joint probability of pre-synaptic and post-synaptic activities, $p_{ij}$, as follows:

$$p_i := (1 - \alpha)\, p_i + \alpha\, \pi_i, \tag{3}$$

$$p_j := (1 - \alpha)\, p_j + \alpha\, \pi_j, \tag{4}$$

$$p_{ij} := (1 - \alpha)\, p_{ij} + \alpha\, \pi_i\, \pi_j, \tag{5}$$

where $\alpha$ is the learning rate. The bias and weight parameters are computed from the $p$-traces as follows:

$$b_j = \log\ p_j, \tag{6}$$

$$w_{ij} = \log \frac{p_{ij}}{p_i p_j}. \tag{7}$$

As a crucial departure from traditional backprop-based DNNs, the learning rule above is local, correlative, and Hebbian, i.e., dependent only on the pre-synaptic and post-synaptic activities. The activity propagation rule (Eq. 1-2) and the learning rule (Eq. 3-7) together constitute the universal BCPNN model, i.e. applicable to both recurrent and feedforward components [24–27].

### 2.4 Network model

The BCPNN model organized as a recurrent network (source layer is the same as the target layer in Eq. 1-7) implements fixed-point attractor dynamics since the weights are symmetric (Eq. 7). Due to the sparse activations and Bayes optimal learning rule, the recurrent-BCPNN network has been shown to have higher storage capacity compared to Hopfield networks when trained on random patterns [28].

Recent work has demonstrated that a feedforward-BCPNN model (source layer encodes data and the target layer is the hidden layer in Eq. 1-7) can perform unsupervised representation learning [16, 17]. This is possible when the model is augmented with structural plasticity mechanisms that learns a sparse set of connections from the input layer to hidden layer. The representations learnt by the model were evaluated on linear classification on various machine learning benchmark datasets and the feedforward BCPNN network was found to favorably compare with many unsupervised brain-like neural network models as well as with multi-layer perceptrons [16, 17].

**Table 1.** List of simulation parameters.

| Symbol | Value | Description |
|---|---|---|
| $H_i$ | 784 | Num. of hypercolumns in input layer |
| $M_i$ | 2 | Num. of minicolumns per hypercolumn in input layer |
| $H_h$ | 100 | Num. of hypercolumns in hidden layer |
| $M_h$ | 100 | Num. of minicolumns per hypercolumn in hidden layer |
| $p_{conn}$ | 10% | Connectivity from input layer to hidden layer |
| $\alpha$ | 0.0001 | Synaptic learning time constant |
| $T$ | 20 | Number of timesteps for iterating recurrent attractor |

We incorporated both recurrent attractor and feedforward components in our network model (Fig. 1A). For this, we modelled a L4+L2/3 hidden layer: L4 units receive feedforward connections from the input layer, and L2/3 units receive input from other L2/3 units (and from the L4 unit within the same minicolumn). We employed the feedforward BCPNN augmented with structural plasticity and used a hidden layer with 100 hypercolumns ($H_h = 100$), each with 100 minicolumns ($M_h = 100$). The recurrent component in the hidden layer was implemented with full connectivity. The network was trained by first learning the feedforward BCPNN component [17], then the feedforward-driven hidden activations were used to train the recurrent weights with the same Hebbian-Bayesian learning. We trained on the 60000 samples of MNIST hand-written digits dataset. For evaluation purposes we used the 10000 samples from the test set and the recurrent attractor was run for $T = 20$ timesteps. Table 1 summarizes all the parameters used in the simulations.

## 3 Results

### 3.1 Feedforward-driven representations are suitable for associative memory

We trained the network on the MNIST dataset and simulated the attractors over the test dataset. To visualize the attractor representation, we back projected (top down) the hidden representations to the input layer using the same weights as feedforward connectivity. For comparison, we also trained a recurrent attractor on the input data. We observed that the recurrent attractors trained on the feedforward-driven representations converged in a few steps, and their reconstruction correspond to prototypical digit images (Fig. 2B). This was in contrast with the severe crosstalk reflected in the behavior of the recurrent attractor network trained directly on the input data (Fig. 2A). The results suggest that the recurrent attractor network operating on the hidden representations enables better associative memory. This is likely because the feedforward-driven hidden activations are orthogonalized, i.e., the representations from the same class are similar and overlapping, while representations from different classes are more distinct. In that case, the activation patterns used for recurrent network learning tend towards "orthogonal" patterns, an ideal condition for associative memory storage to avoid interference between attractor states.

To test this more systematically, we computed the pair-wise similarity matrix on *i)* the input data (Fig. 2C), *ii)* activations after running the recurrent attractor on the input data (Fig. 2D), *iii)* feedforward-driven hidden representations (Fig. 2E), and iv) recurrent attractor hidden representations (Fig. 2F). We used the cosine similarity score as the metric and sorted the rows and columns by the sample class (label). Ideally, the similarity for samples from the same class should be close to one (high overlap between representations), and close to zero for samples from different classes (minimal overlap between representations). We quantified the overall degree of orthogonality by calculating a ratio of mean similarity of samples within the same class and mean similarity across all samples (higher ratio implies more orthogonal representations). The similarity matrix for the input data showed class ambiguity (the orthogonality ratio was 1.37; Fig. 2C) as samples were similar to other samples within and outside the class, and the cosine score was roughly around 0.5 for most sample pairs indicating considerable overlap. The similarity matrix of attractors when run on the input data was largely disordered (the orthogonality ratio was 1.47; Fig. 2D), indicating heavy crosstalk from learning the input data. The similarity matrix for the feedforward-driven hidden representations were markedly different (the orthogonality ratio was 3.79; Fig. 2E), with most of the cosine score close to zero (because the activations are sparse), and the few non-zero scores occurred within the same class (small square patches along the diagonal in Fig. 2E). The feedforward network can be said to untangle the representations, i.e., those from the same class are made similar and overlapping, while representations from different classes are pushed far-apart. When the recurrent attractor layer was trained on these feedforward representations, the attractors learnt were even more orthogonal (the orthogonality ratio was 7.54; Fig. 2F) since many samples from the same class settled on the same prototype attractor. Hence, the feedforward-driven hidden activations of the BCPNN model are sparse, distributed, high-dimensional and orthogonal – in other words, having all the attributes ideal for associative memory.

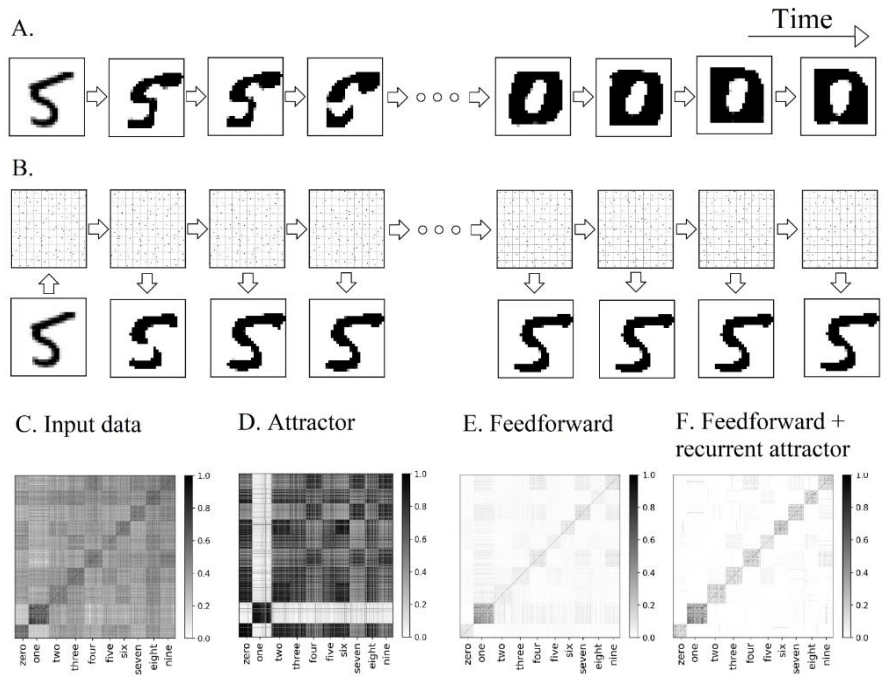

**Fig. 2. Feedforward-driven representations are suitable for associative memory. A.** Recurrent attractor activations of the network trained on the input data and tested on the image of digit "5" show convergence to a spurious pattern because of severe crosstalk between stored memories. **B.** Recurrent attractors trained on the feedforward-driven representations (boxes in the top row with dots representing active minicolumns) and tested on the same digit "5" converged in a few steps to one of the attractor prototypes, as reflected by the input reconstruction (boxes in the bottom row). **C, D, E, F.** Pair-wise similarity matrix for the input data (C), attrac-

tor activations when trained on the input data (D), feedforward-driven representations (E), attractor activations when trained on the feedforward-driven representations (F).

### 3.2   Associative memory extracts prototypes from training data

Prototype extraction facilitates identification of representative patterns for subsets of data samples. Here we examine if the attractors learnt by the recurrent network component can serve as prototypes even if the network is not explicitly trained on any prototypical data. We expect that the attractor prototypes account for semantically similar data, for example, data samples sharing many common features or images from the same class.

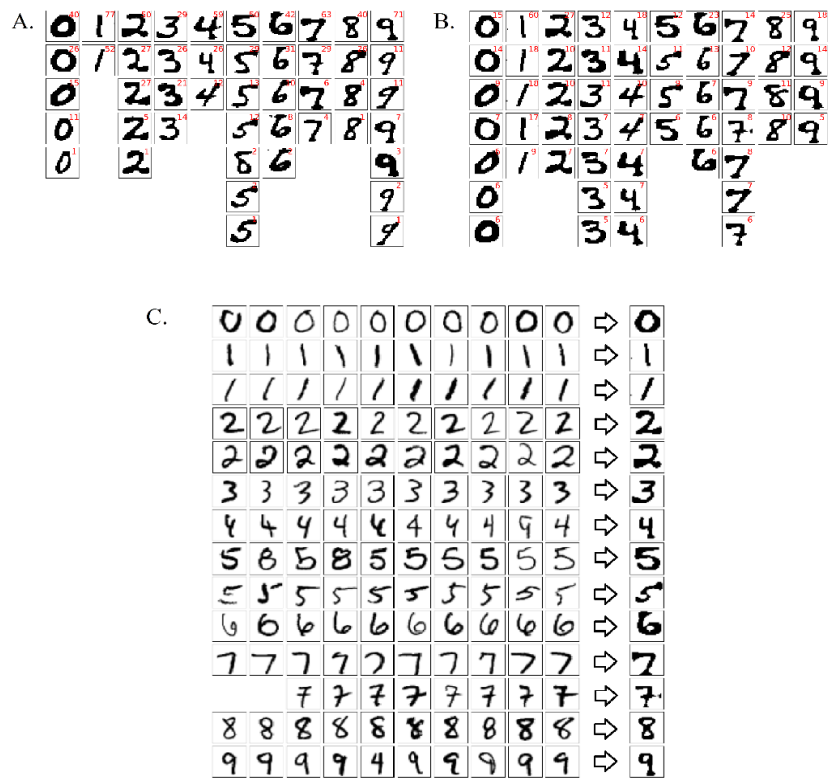

**Fig. 3. Prototypes extraction. A**, **B**: Top-down reconstructions of the unique prototypes extracted by the recurrent attractor network for 1000 test samples. Prototype patterns were considered unique if the cosine similarity was more than a threshold value set to 0.5 (A) or 0.9 (B). The red number shown in the top right corner of each image indicates the number of input samples (out of 1000) that converged to the prototype. **C.** Randomly selected input samples that converged to the same prototype attractor (input reconstruction in the last column).

To test this, following the training process described earlier we ran our full network model (feedforward propagation followed by recurrent attractor for $T$ steps) with 1000 samples randomly selected from the unseen test subset of the MNIST da-taset to obtain the final attractor pattern for each sample. We examined these attractor patterns to assess if they represented unique prototypes. Since we used real-valued activation patterns (from Eq. 2), we could not directly check if patterns are identical to one another. Instead, we computed the cosine similarity between attractor patterns (same as Fig. 2D) and used a threshold to determine if they were similar enough to be considered the same prototype. As a result, the number of "unique" prototypes depends on the threshold. For illustrative purposes we used two different threshold values and computed the prototypes as well as their corresponding reconstructions. For the threshold of 0.5 we found 46 prototypes (shown in Fig. 3A) and for $0.9 - 248$ prototypes (prototypes with more than 5 input samples are shown in Fig. 3B). In Figs. 3A-B the prototype reconstructions are manually grouped into columns by their class and sorted in rows by the number of distinct samples that converge to the same prototype (descending order). We can observe that in both cases the prototypes account for all the classes and different styles of writing: upright and slanted ones, dashed and undashed seven, etc.

Fig. 3C shows in rows groups of randomly selected input samples that converged to the same attractor prototypes (14 picked from the prototypes obtained with the similarity threshold of 0.9). The prototypes mostly encode some salient features of the data, which reflects grouping according to their class. It is worth noting that the network identified in an unsupervised manner more than a single prototype per class. For instance, there are different prototypes for upright one and slanted one ($2^{nd}$ and $3^{rd}$ row, Fig. 3C), twos without and with a knot ($4^{th}$ and $5^{th}$ row), undashed seven and dashed seven ($11^{th}$ and $12^{th}$ row). Interestingly, there are some prototypes that include samples with different classes, for example a nine without a closed knot converges to the prototype four ($7^{th}$ row, $9^{th}$ column), a wide eight with a slim base converges to the prototype nine ($15^{th}$ row, $8^{th}$ column). Still, these examples are perceptually intuitive irrespective of the ground truth labelling.

### 3.3  Recurrence makes representation robust to data distortions

We next examined if the recurrent attractor network added value to the feedforward-driven representations when tested on severely distorted samples. For this we first created a distorted version of the MNIST dataset (examples shown in Fig. 4A) following the work of George *et al.*, (2017). In particular, we introduced nine different distortions (Fig. 4A rows) and controlled the level of distortion with a distortion level parameter ranging from 0.1 (minimum distortion) to 1.0 (maximum distortion) in step of 0.1 (Fig. 4A columns). In total, we created the distorted dataset with 900 samples (from the test set). Then we ran the network, trained beforehand on the original undistorted MNIST training dataset as before, with these distorted samples as inputs to comparatively study the recurrent network attractor activations and their corresponding input reconstructions. Except for high distortion levels (0.5 or higher), the attractor network was very robust to the distortions, and the reconstructed images showed that most of the distortions were removed upon attractor convergence. Fig 4B shows examples of two digits under all distortion types (distortion level of 0.3) and the reconstructions of the corresponding attractor activations.

To quantify the network's robustness to input distortions in the pattern recognition context, we used a 10-class linear classifier (one-vs-rest), which was trained on the feedforward-driven activations (on the original undistorted MNIST training set). We then compared the classification accuracy from five random trials obtained for recurrent attractor representations and the feedforward-driven hidden representations on the distorted MNIST dataset (Fig. 4C and 4D). We found that the recurrent attractor representations performed better compared to feedforward-driven representations on all distortion levels greater than 0.1 (Fig. 4C). We also examined the performance across the different distortion types and the recurrent attractor representations turned out more resilient in most cases (Fig. 4D). Further investigation is needed to understand why feedforward-driven representations scored higher on three distortion types: "Grid", "Deletion", and "Open". Both feedforward-driven and recurrent attractor representations result from unsupervised learning, so the distortion resistance provided by the recurrent component is achieved without any access to data labels.

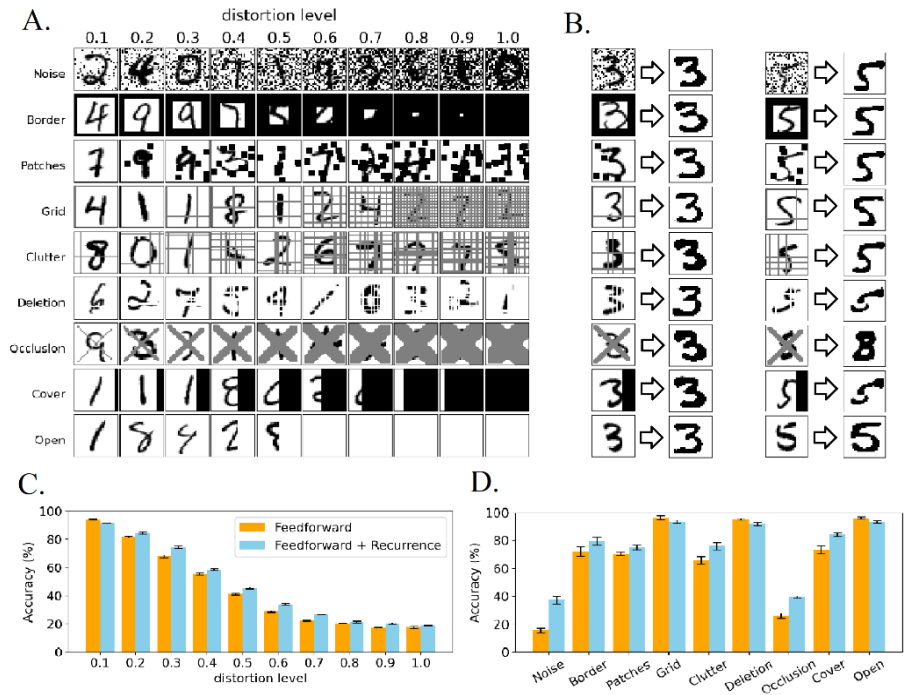

**Fig. 4. Robustness to distortion A.** Samples from the distorted MNIST dataset, with 9 distortion types (rows) and varying distortion levels (columns). **B.** Examples of digits "3" (left) and "5" (right) under all distortion types (distortion level = 0.3), and the input reconstruction after convergence of the recurrent layer. For most cases, the attractor reaches the prototype pattern while removing all distortions. **C.** Classification performance for different distortion levels comparing feedforward-driven and recurrent attractor representations. **D.** Classification performance for all distortion types with distortion level at 0.3.

## 4   Related work

Feedforward networks are currently the dominant approach for building perceptual systems [11]. Although feedforward networks make use of some salient aspects of the brain, such as the hierarchical processing stages found in the cortical sensory pathway [30, 31], there are reasons to believe this is not sufficient. Given the abundance of recurrent structures in the brain, their function in perception and biological vision has been studied extensively over the years. There is considerable evidence that recurrence is responsible for many important visual computations such as pattern completion [32], figure-ground segmentation [33], occlusion reasoning [34, 35] and contour integration [36, 37]. Specifically, object recognition tasks under challenging conditions (e.g. occlusions, distortions) are thought to engage recurrent processes [38].

Early computational models incorporated recurrent interactions between units in the hidden layer as a mechanism to encourage competition. Such competition creates mutually inhibiting neurons, and allows feedforward learning of localist representations that clusters the input data [39–41]. More recently, the computational potential of recurrence has come into focus with the motivation to improve the robustness of DNNs. Feedforward DNNs are observed to be fragile to distortions as well as "adversarial" examples (minor distortions imperceptible to humans) [29, 42, 43]. Thus, incorporating it into feedforward networks has been shown to enhance their recognition performance, especially under challenging conditions [2, 44]. Augmenting DNNs with a recurrent attractor network component (on the top feature layer) significantly improved the classification performance on images with occlusion and provided a closer fit to human recognition performance [32]. Work on probabilistic models (as opposed to neural networks) has also shown that recurrent interactions implement contour integration and occlusion reasoning, thereby enabling models to solve object recognition tasks where classical DNNs fail [29, 43].

## 5   Conclusions and outlook

In this work, we demonstrated a tight integration of a recurrent attractor network component and a feedforward component within the BCPNN framework. We showed that the resulting network (1) forms associative memory by learning attractors in the space of hidden representations, (2) performs prototype extraction and (3) renders the representations more robust to distortions compared to the hidden representations obtained with a feedforward network.

The workings of the network can be summarized as follows: the feedforward network component learns in an unsupervised manner a set of salient features underlying the data and forms sparse distributed representations, while the recurrent attractor network component acts on the feedforward-driven hidden representation and performs pattern reconstruction by integrating information across the distributed representation and forming a coherent whole in the spirit of Gestalt perception. This design principle comes directly from the circuitry found in the neocortex: within each cortical minicolumn feedforward connections arrive at excitatory neurons located in the granular layer 4, L4 neurons project to neurons in the superficial layer 2/3, L2/3 neurons make extensive recurrent connectivity with L2/3 neurons in other minicolumns before projecting to L4 neurons of higher cortical areas. Such circuit motifs are considered canonical in the neocortex [19, 20] and offer powerful computational primitives that, we believe, need to be understood for building intelligent computer perception systems as well as for exploring the perceptual and cognitive functions of the cortex.

Prior work has shown that DNNs perform poorly on distorted images (such as those used in this work) despite having high accuracy on clean images [29]. As the next step, we plan to extensively compare our model with other state-of-the-art methods [29, 32] in such challenging pattern recognition scenarios. Other outstanding problems include among others a seamless functional integration of top-down back-projections in the spirit of predictive coding (which we utilized in this work only for visualization purposes) [45, 46]. Future work will also focus on how multilayer deep network formed by stacking such L4-L2/3 modules can be built and trained most efficiently. Extending from the static data domain, the synaptic traces at the core of the BCPNN model can also be configured to perform temporal association and sequence learning and generation [47, 48].

The research direction we pursued here integrates various architectural details from neurobiology, especially from the mammalian neocortex, into an abstract cortical model, which is then operationalized as a system of functionally powerful unsupervised learning algorithms. The resulting information processing machinery holds promise for creating novel intelligent brain-like computing systems with a broad spectrum of technical applications.